\documentclass[letterpaper, 10 pt, conference]{ieeeconf}  
\IEEEoverridecommandlockouts 
\usepackage{cite}
\usepackage{hyperref}
\usepackage[svgnames]{xcolor}
\hypersetup{
    colorlinks,
    linkcolor={red!50!black},
    citecolor={blue!50!black},
    urlcolor={blue!80!black}
}
\usepackage{amsmath,amssymb,amsfonts}
\usepackage{algorithmic}
\usepackage{graphicx}
\usepackage{textcomp}
\usepackage{booktabs}
\usepackage{mdframed}
\usepackage[many]{tcolorbox}
\usepackage[para,online,flushleft]{threeparttable}
\def\BibTeX{{\rm B\kern-.05em{\sc i\kern-.025em b}\kern-.08emS
    T\kern-.1667em\lower.7ex\hbox{E}\kern-.125emX}}

\usepackage{inconsolata}
\usepackage[T1]{fontenc}
\usepackage{listings}

\usepackage{url}

\definecolor{light-gray}{rgb}{0.8, 0.8, 0.8}
\definecolor{prompt-example}{HTML}{003366}
\definecolor{prompt-result}{HTML}{006633}
\definecolor{system}{HTML}{333333}
\definecolor{highlight-result}{HTML}{ECFFDC}
\definecolor{highlight-example}{HTML}{ADD8E6}

\definecolor{diffincl}{named}{Green}
\definecolor{diffrem}{named}{OrangeRed}

\lstdefinelanguage{diff}{
    basicstyle=\scriptsize\ttfamily,
    morecomment=[f][\color{diffincl}]{+\ },
    morecomment=[f][\color{diffrem}]{-\ },
}

\begin{document}


\title{\LARGE \bf
BTGenBot: Behavior Tree Generation for Robotic Tasks with Lightweight LLMs
}

\author{Riccardo Andrea Izzo$^{\dag}$, Gianluca Bardaro$^{\dag}$, and Matteo Matteucci$^{\dag}$
\thanks{$^{\dag}$All authors are with the Department of Electronics, Information, and Bioengineering, Politecnico di Milano, Milan, Italy. E-mail:
        {\tt\small name.surname@polimi.it}}%
}


\maketitle

\begin{abstract}
This paper presents a novel approach to generating behavior trees for robots using lightweight large language models (LLMs) with a maximum of 7 billion parameters. The study demonstrates that it is possible to achieve satisfying results with compact LLMs when fine-tuned on a specific dataset. The key contributions of this research include the creation of a fine-tuning dataset based on existing behavior trees using GPT-3.5 and a comprehensive comparison of multiple LLMs (namely llama2, llama-chat, and code-llama) across nine distinct tasks. To be thorough, we evaluated the generated behavior trees using static syntactical analysis, a validation system, a simulated environment, and a real robot. Furthermore, this work opens the possibility of deploying such solutions directly on the robot, enhancing its practical applicability. Findings from this study demonstrate the potential of LLMs with a limited number of parameters in generating effective and efficient robot behaviors.
\end{abstract}


\section{Introduction}
In recent years, robots have become an integral part of our everyday lives, permeating various sectors such as logistics, manufacturing, and healthcare.
The increasing prevalence of robots in these diverse fields shows the necessity to go beyond a simple set of functionalities and achieve adaptable, flexible, and effective task planning. This is necessary to tackle the challenges of modern robotics, such as dynamic and unstructured surroundings, unpredictable situations, and advanced interaction with humans and the environment.
To achieve this, various representations have been proposed to describe robot tasks~\cite{guo2023recent} and multiple planning languages~\cite{dragule2021languages} have been designed to let experts write complete and descriptive definitions of high-level tasks.

An example of these representations is behavior trees (BTs). Behavior trees have gained significant traction in robotics, offering a structured and scalable approach to managing complex robot behaviors. Originating from the video game industry, BTs have found relevance in robotics due to their ability to handle high-level decision-making and control in dynamic environments. A key player in this transition is the BehaviorTree.CPP library\footnote{https://www.behaviortree.dev/}. This library has become the de facto standard in the Robot Operating System 2 (ROS2) ecosystem, largely due to its inclusion in Navigation2~\cite{macenski2020marathon}. This integration has facilitated the development of sophisticated navigation behaviors, contributing to the broader adoption and standardization of BTs in robotics software design.

However, even after a long history of efforts, there is no definitive solution to equip robots with flexible and adaptable task planning, since it requires complex knowledge, such as the robot's actions, the dynamic nature of the environment, object relations, and affordances.

Recently, large language models (LLM) trained on large amounts of data have emerged as a promising tool to manage this challenge and deal with the complex knowledge required for task planning. These models have demonstrated remarkable success in robot planning from natural language instructions. Additionally, an LLM can represent commonsense priors (e.g., visit the colder locations first) and comprehend spatial relationships (e.g., move behind the sofa). However, such large models present a major limitation: their substantial hardware resource requirements prevent their direct deployment on robots.

This work explores an alternative approach, focusing on compact large language models, specifically those with 7 billion parameters. We posit that these smaller and less resource-intensive models can generate behavior trees for robots with comparable effectiveness to their larger counterparts. Our main contributions are: (i) a dataset of 600 behavior trees paired with a natural language description of their tasks, (ii) a system to evaluate and validate the performance of the generated BT over nine different tasks including navigation and manipulation, and (iii) a comparative analysis of three different LLM and their fine-tuned versions. All the material used in this work, including the dataset, the complete prompts of both evaluation phases, and the source code of the validator are available at \url{https://github.com/AIRLab-POLIMI/BTGenBot}.

\section{Related work}
A pre-trained foundation model (PFM) is a complex neural network-based model trained on extensive open-source datasets, boasting billions of parameters~\cite{zhou2023comprehensive}. This robust architecture serves as a versatile backbone that can be further fine-tuned for diverse tasks across various domains.
Recently, several foundation models specifically trained on massive text datasets (i.e., large language models) have been proposed such as the most recent one GPT-4~\cite{achiam2023gpt} by OpenAI or LLaMA by Meta AI~\cite{touvron2023llama}.
In this work, we rely on LLaMA released by Meta AI, a family of foundation models trained on an open-source dataset with trillions of tokens. In particular, we employed LLaMA-2~\cite{touvron2023llama2}, trained on 2 trillion tokens, and with a context length of 4000 tokens, double the length of its predecessor. Three model sizes are available: 7, 13, and 70 billion parameters.

\textbf{Behavior tree generation}. An example of behavior tree generation using LLM is the work done in~\cite{lykov2023llm}. In this paper, the authors exploit a transformer-based LLM fine-tuned from the Stanford Alpaca 7B model~\cite{taori2023alpaca} to generate behavior trees from a text description. They further fine-tuned this model with natural language descriptions generated using \textit{text-davinci-003} from behavior trees created in the previous step. In summary, they use two different LLMs to generate both the behavior trees and the task description. While interesting in proving that LLMs can be used to generate BTs, this approach has several limitations. Fine-tuning an LLM with a dataset generated by the LLM itself causes a data bias that is only partially mitigated by using an auxiliary LLM to generate the descriptions. Moreover, self-generated data can be noisy and, in this configuration, errors propagate from the data generation phase to the training phase. Differently from the solution proposed in~\cite{lykov2023llm}, we create our instruction-following dataset by collecting behavior trees from open-source robotics projects. This allows us to fine-tune our model with BTs already tested and effectively used in real projects. Moreover, we use GPT-3.5 only to generate descriptions for each corresponding behavior tree.

Finally, concerning validating the above approach, the only metric used is the visual evaluation by a group of human experts. The evaluators were presented with a pair of behavior trees and had to indicate which one was generated.
This evaluation, while interesting and useful to asses the general capabilities of the model, is rather limited. This method does not consider the correctness from a syntactical and semantic point of view, or any performance metric, such as the generation time and the hardware resources used. In contrast, our approach aims at creating fully functional behavior trees to be used in real-world applications. Therefore, we tested the generated outputs with the metrics described in Section~\ref{methods}. 

\textbf{Code-based planning}. Another approach to defining robot behaviors using LLM is to generate executable code. An excellent example is Code as Policies~\cite{liang2023code}. In this work, the authors release a robot-centric formulation of language model generated programs (i.e., LMPs), based on hierarchical code generation prompting, that allows the generation of new policy code via recursively defining undefined functions.
LMPs use few-shot prompting to generate different subprograms and can be generated hierarchically with chain-of-thought prompting. Even if their approach requires no additional training, their system is based on GPT-3 with 175B parameters. Moreover, they use open-vocabulary object detection models like ViLD~\cite{gu2021open} and MDETR~\cite{kamath2021mdetr} to compose perception-to-control feedback logic. Another work that relies on prompt engineering to create a Pythonic planner is ProgPrompt~\cite{singh2023progprompt}. They introduce a programmatic LLM prompt structure that enables plan generation functional across situated environments and robot capabilities. Their fine-tuned version of GPT-2 can generate plausible action plans in the context of robotic task planning, this is achieved by providing the model with a Pythonic program-like specification of the the available actions and objects in an environment and executable example programs. In this way, they introduce situated awareness in LLM-based robot task planning via prompting.

Both works described in~\cite{liang2023code} and~\cite{singh2023progprompt} benefit from LLMs with over 100B parameters, while our approach instead leverages smaller models up to only 7B parameters. This allows our method to be replicable locally on consumer hardware without relying on larger, more complex models available via APIs. Additionally, the use of behavior trees gives more flexibility in terms of expressive ability, represents a plug-and-play solution as in~\cite{ogren2022behavior}, and is commonly used in recent open-source robotics projects as demonstrated by~\cite{iovino2022survey}.

\begin{figure}[t]
\centering
\includegraphics[width=0.9\columnwidth]{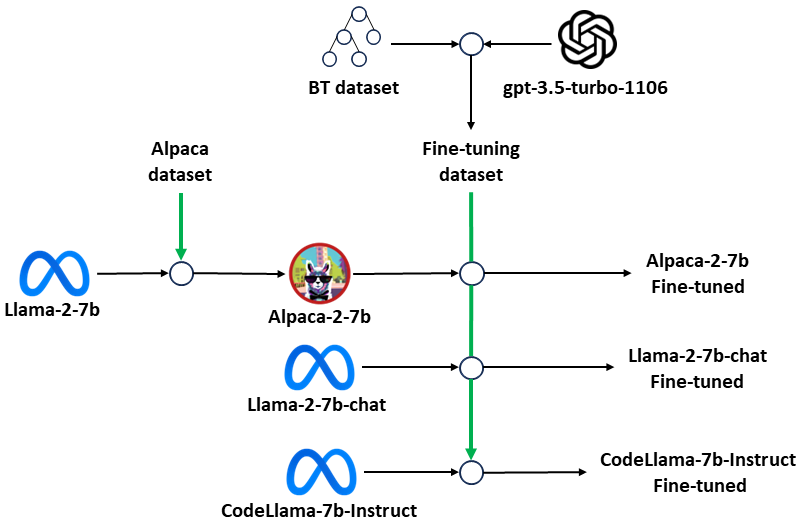}
\caption{Fine-Tuning Process}
\label{fig:tuning}
\end{figure}

\section{Method}
\label{methods}
For the experiments in this work, we considered the state-of-the-art open-source large language models provided by Meta as reference models. We selected \textit{Llama-2-7b}~\cite{touvron2023llama2}, the foundation model, \textit{Llama-2-7b-chat}~\cite{touvron2023llama2}, designed for conversational tasks, and \textit{CodeLlama-7b-Instruct}~\cite{roziere2023code}, with a focus on code-generation. These models are based on the transformer architecture pioneered in~\cite{vaswani2017attention}. Following the approach proposed in~\cite{lykov2023llm}, we fine-tuned the foundation model \textit{Llama-2-7b} with the Alpaca dataset, with the procedure reported in~\cite{taori2023alpaca} using the Alpaca-LoRA repository~\cite{tloen2023lora}. For the remainder of this paper, we will refer to this fine-tuned version as Alpaca.
Additional fine-tuning of these models allows for customization to meet specific requirements, such as domain specialization (e.g., the generation of behavior trees). A summary of the various fine-tuning steps is presented in Figure~\ref{fig:tuning}, and a more detailed description is provided in Section~\ref{sec:tuning}.

\subsection{Dataset Format}
\label{sub_dataset_format}
Each entry of our instruction-following dataset has three parts: ``instruction'', ``input'' and ``output''. While the original work of~\cite{taori2023alpaca} states that ``input'' is optional, in our case, we combine the XML version of the behavior tree provided in ``output'' with the description of the task in the ``input'' to create a definition that matches the prompts used later during inference.

\begin{mdframed}[linecolor=light-gray]
\scriptsize{\textsf{\color{system}You will be provided a summary of a task performed by a robot, and your objective is to express this task as a behavior tree in XML format.}}
\vspace{0.6em}

    \noindent \scriptsize{\textsf{\color{prompt-example}The behavior tree is a simple sequential task for a robot. It first instructs the robot to move to a specific point (GoPoint) and then to interact with a particular object (GoObject). The robot will execute these tasks in sequence, moving to the specified destination before interacting with the designated object.}}
    \vspace{-0.6em}
    \begin{mdframed}[backgroundcolor=highlight-example, innertopmargin=-4pt, innerbottommargin=-4pt, innerleftmargin=2pt, innerrightmargin=2pt]
    \lstset{
        language=xml,
        stringstyle=\color{red},
        basicstyle=\scriptsize\ttfamily,
    }
\begin{lstlisting}
<root main_tree_to_execute = "MainTree" >
  <BehaviorTree ID="MainTree">
  <Sequence name="root_sequence">
    <GoPoint goal="{destination}"/>
    <GoObject target="{object}"/>
  </Sequence>
  </BehaviorTree>
</root>
\end{lstlisting}
    \end{mdframed}
\end{mdframed}

This is an example of an entry in the dataset. In gray, the ``instruction'' element is common to all the samples in the dataset, and it is fixed and immutable. It is followed by the ``input'' element, in blue, where a description in natural language is provided of the behavior tree. Finally, the ``output'' element represents the behavior tree in XML format.
All the behavior trees used in the dataset are sourced via the work of~\cite{ghzouli2023trees}. In this work, the authors provide a collection of roughly 600 behavior trees collected from various open-source projects in the field of robotics. This dataset is particularly valuable because compatible with BehaviorTree.CPP, the de facto standard of ROS2, and the same format used by~\cite{lykov2023llm}. Furthermore, what distinguishes this dataset is the quality of the behavior trees, which have been developed for actual applications and tested on robots and thus have already been validated in real-world scenarios. The natural language description (i.e., the ``input'' element) of each dataset entry has been generated automatically using GPT-3.5, with a procedure described in the next section.

\subsection{Dataset Generation}
Recent studies such as~\cite{zhao2023survey} and~\cite{bubeck2023sparks} highlight the exceptional capabilities of modern LLM, notably GPT, in several domains including question answering, text generation, and code generation.
GPT models are suitable for text generation tasks~\cite{zhao2023survey} and can be exploited to generate a description of the provided input.
Considering that, we used OpenAI available APIs to complete our instruction-following dataset. In particular, the ``input'' element (i.e., the description of the behavior tree) has been generated using the \textit{gpt-3.5-turbo} model with default parameters and a context length of 2048 tokens. We prompted the model to generate a natural language description of a behavior tree in XML format received as input. The one-shot prompt used with \textit{gpt-3.5-turbo} model to generate the description of the provided behavior tree is the same discussed in~\cite{taori2023alpaca} and it is composed of the following elements: ``system'', ``user'', ``assistant'' and again ``user''. The ``system'' element is shared between all prompts and represents the general context given to the model: ``You will be provided a behavior tree in XML format, and your task is to summarize the task performed by this behavior tree''. The above prompt represents only the starting point, additional information is given including the maximum number of words and the required compatibility with BehaviorTree.CPP library and the fact that the description must represent an overall summary of the task clearly described in natural language. This method ultimately leverages the inherent linguistic capabilities of GPT models to create a description of the behavior tree. To assess the quality of the generated descriptions, we sampled a subset of the results (i.e., roughly ten behavior trees) and evaluated visually how well the description matched the input BT. The result of our evaluation was positive, therefore we proceeded to generate the whole dataset using the procedure described before.

\subsection{Fine-Tuning Process}
\label{sec:tuning}
The models selected for the fine-tuning process are \textit{Llama2-7b}, \textit{LlamaChat}, and \textit{CodeLlama-Instruct-7b}. The fine-tuning process can be structured in two steps, the first involves the \textit{Llama2-7b} model that has been fine-tuned with the original Alpaca dataset as reported in~\cite{hu2021lora}. This step prepared the base model for instruction-following tasks, differently from the other models already fine-tuned for this task. The second step prepared the models for the generation of the behavior trees. We further fine-tuned the models to perform this task using the dataset described in \ref{sub_dataset_format}. LLMs fine-tuning is a computationally intensive task, therefore, we employed the Parameter-Efficient-Fine-Tuning (PEFT) approach, in particular, Low-Rank Adaptation (LoRA) described in detail in~\cite{hu2021lora}. As reported in~\cite{pu2023empirical}, this is particularly useful to avoid retraining the entire model from scratch. This way, we can keep the model frozen and add an adapter at the end of the model consisting of a few learnable parameters, and layers. 

 Most LoRA parameters were set at their default value, except for the target modules. In the work~\cite{hu2021lora}, it is noted that the transformer architecture has four weight matrices in the self-attention module (\textit{$W_{q}$}, \textit{$W_{k}$}, \textit{$W_{v}$}, \textit{$W_{o}$}). To enhance generalization capabilities with our limited dataset and to optimize performance, it was necessary to unfreeze the MLP layers. We expanded the corresponding initial set of weight matrices [q\_proj, k\_proj, v\_proj, o\_proj] by adding the additional weight matrices of the MLP module, these include [gate\_proj, up\_proj, down\_proj]. This approach led to a more robust and effective training process.

 The fine-tuning process was accomplished with two NVIDIA RTX Quadro 6000 with a total of 48GB of video memory. All the basic hyperparameters have been used except for a batch size of 256 and a micro-batch size of 4, we used a learning rate of 3e-4 and a validation set size of 5\%. Given that our dataset contains less than a thousand samples, we increased the number of epochs to achieve satisfying results.

\section{Evaluation}
To assess and compare the performance of the three base models (i.e., Alpaca, LlamaChat, and CodeLlama) and our fine-tuned versions, we conducted different tests using nine task descriptions of behavior trees.

\subsection{Task definition}
All three models were tested, with the input being the provided description of the behavior tree. The expected output is a behavior tree in XML format, compatible with BehaviorTree.CPP library, which reflects the provided description.
The tasks used are the following:
\begin{enumerate}
\item \textbf{Navigation.} The robot is tasked to reach a series of locations provided as coordinates in a specific order.
    \item \textbf{Navigation with priority.} The system receives a list of locations and a list of corresponding readings (e.g., temperature). The robot must visit all locations prioritizing those with a reading above a given threshold. 
\item \textbf{Navigation with fallback.} The robot must navigate through a series of waypoints. During the navigation, a waypoint may become unreachable. In this case, the destination must be skipped and the robot should proceed to the next waypoint.
\item \textbf{Navigation with arm activity.} An extension of the Navigation task,  at each location the robot activates the on-board manipulator.
\item \textbf{Exploration.} In this situation, the robot navigates continuously. Periodically, the robot receives a new location and checks if the exploration routine is completed.
\item \textbf{Manipulator exploration.} A task performed by a manipulator. The system cycles through multiple joint configurations until a target object is found. When found, the manipulator approaches the target.
\item \textbf{Active vision and picking.} A robotic arm observes an object, estimates a grasping position, and performs a "pick and place" routine.
\item \textbf{Material processing.} A robot manipulator triggers material processing by pressing buttons in the correct sequence. The robot is also in charge of evaluating the status of the processing.
\item \textbf{Multi-station assembly.} A mobile manipulator moves between multiple stations to collect components and assemble an object.
\end{enumerate}

\subsection{Evaluation approach}
The evaluation has been accomplished in two phases: first a preliminary selection among the models, and then a validation of the models and the generated behavior trees with more rigorous metrics. In the first phase, we evaluated the models on a subset of simpler tasks (i.e., tasks 1 to 7). In the second phase, where the general capabilities of the models are already been assessed, all nine tasks are considered. 
The metrics considered in the first evaluation phase are:
\begin{itemize}
    \item \textbf{Time}: assess the time performance of the models, this is crucial when the entire pipeline, from the LLM to the actual robot, is considered.
    \item \textbf{Syntactic Correctness}: evaluating the syntactic accuracy using Groot2, an IDE to create and debug behavior trees provided by BehaviorTree.CPP. This allows us to check the syntactic correctness of the XML schema and the overall tree.
    \item \textbf{Semantic Evaluation}: without considering the syntactic correctness, we evaluate the semantic accuracy. This means the ability of the LLM to generate a solution that understands and solves the problem at hand. A group of human experts performs this evaluation.
\end{itemize}


\begin{figure}[t]
\centering
\includegraphics[width=0.6\columnwidth]{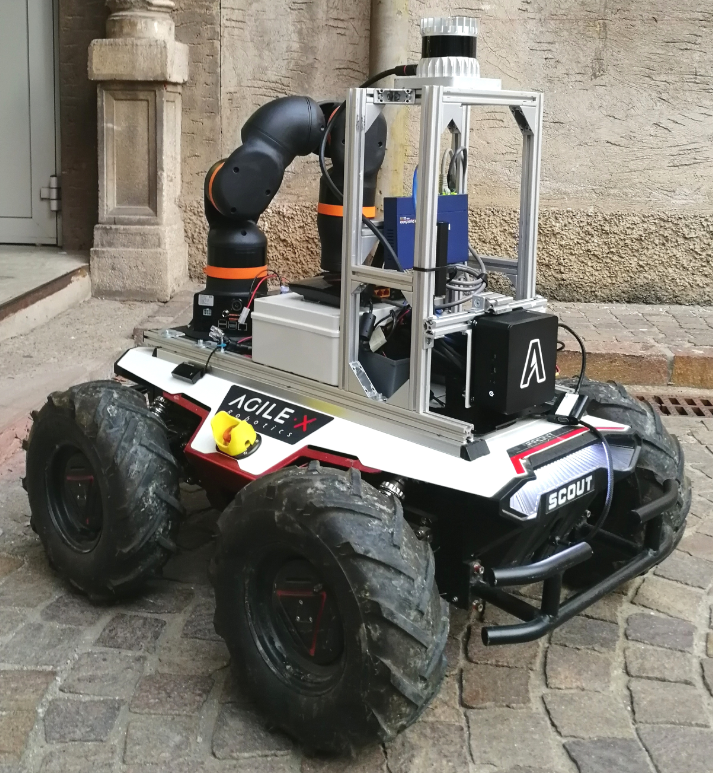}
\caption{Robot used to test the generated behavior trees}
\label{robot}
\end{figure}

Moving on to the second evaluation phase, we considered the following metrics:
\begin{itemize}
    \item \textbf{BT Correctness}: validating the behavior trees using a custom-developed behavior tree validator
    \item \textbf{Simulation robot}: verifying the successful execution of the task on a simulated environment, in particular, the navigation task has been tested on the TurtleBot3. 
    \item \textbf{Real robot}: verifying the successful execution of the task on the physical robot shown in Figure~\ref{robot}.
\end{itemize}


\begin{table}[t]
\caption{Performance reference}
\label{validation-table-gpt}
\centering
\begin{tabular}{@{}llll@{}}
\toprule
       & GPT-4o & Gemini & Llama2 13B \\ \midrule
Task 1 & \checkmark & \checkmark & \checkmark \\
Task 2 & \checkmark &  &  \\
Task 3 & \checkmark & \checkmark &  \\
Task 4 & \checkmark & \checkmark & \checkmark \\
Task 5 & \checkmark &  &  \\
Task 6 & \checkmark &  &  \\
Task 7 &  &  &  \\
Task 8 &  & \checkmark &  \\
Task 9 & \checkmark &  &  \\ \bottomrule      
\end{tabular}
\end{table}

\begin{table}[t]
\caption{Syntactic evaluation (phase 1)}
\label{syntax-table}
\centering
\begin{tabular}{@{}llll@{}}
\toprule
& Alpaca & LlamaChat & CodeLlama    \\ \midrule
Base, Zero-Shot & 0\% & 0\% & 28\% \\
Fine Tuned, Zero-Shot & 14\% & 71\% & 86\% \\ \midrule
Base, One-Shot & 71\% & 86\% & 100\% \\
Fine Tuned, One-Shot & 86\% & 86\% & 86\% \\\bottomrule
\end{tabular}
\end{table}

\begin{table}[t]
\caption{Semantic evaluation (phase 1)}
\label{semantic-table}
\centering
\begin{tabular}{@{}llll@{}}
\toprule
& Alpaca & LlamaChat & CodeLlama    \\ \midrule
Base, Zero-Shot & 14\% & 14\% & 71\% \\
Fine Tuned, Zero-Shot & 28\% & 57\% & 100\% \\ \midrule
Base, One-Shot & 28\% & 57\% & 71\% \\
Fine Tuned, One-Shot & 14\% & 14\% & 57\% \\\bottomrule
\end{tabular}
\end{table}

\subsection{Performance reference}
To provide a frame of reference for the performance achieved by our approach, we tested our nine tasks against state-of-the-art LLMs. We selected GPT-4o, Gemini, and Llama 2 13B Chat. Table~\ref{validation-table-gpt} shows the results. The values reported in the table are obtained using the same one-shot prompts used in our evaluations. The generated behavior trees are validated with our custom validator without any post-processing. GPT-4o achieves the best performance, it can generate a correct solution for seven out of nine tasks. Gemini can provide a working solution only in four tasks. While, as expected, behavior trees generated by Llama 2 13B Chat are correct only in tasks 1 and 4, which are the less challenging. This simple evaluation shows that even larger LLMs may struggle to generate syntactically and semantically correct behavior trees without an accurate prompt or fine-tuning. In particular, in some cases, it is enough to fix small syntactical errors (e.g., correct a Decorator name) to achieve a correct solution. 

\subsection{Preliminary selection}

Since this approach is intended to be used directly on the robot, we compare the inference times of each model when running on a Jetson AGX Orin 64GB. The inference time ranges from a few seconds for simpler tasks to a few minutes when considering more complex prompts. This pattern is confirmed between zero-shot and one-shot prompts, where the longer prompt (i.e., one-shot) has a longer inference time. When discussing inference time, it is worth noting that in some cases, Alpaca fails to terminate the generation process. 

Regarding syntactic correctness, we define a behavior tree as syntactically correct if successfully recognized by Groot2. 
All the outputs of three base models and the corresponding fine-tuned version with our dataset, both with zero-shot and one-shot prompting, have been evaluated using Groot2. The results of this evaluation are summarised in Table~\ref{syntax-table}. Base models with a zero-shot prompt achieve poor results, only CodeLlama can generate a syntactically correct behavior tree in a few cases. Adding examples in the prompt (i.e., one-shot) significantly boosts performance, with CodeLlama creating a syntactically correct behavior tree for each task.  A similar result is achieved for fine-tuned models. In this case, even with a zero-shot prompt, all models can generate syntactically correct behavior trees for most tasks. Surprisingly, providing an example does not drastically increase accuracy as before, with the extreme case of CodeLlama achieving a worse result than before. This result shows how impactful the choice of examples can be on the final output.

Table~\ref{semantic-table} summarises the analysis done by human experts on the semantics of the behavior tree. In this context, by semantic, we mean the ability of the system to interpret the task defined in the prompt and create a solution in an XML format that represents the correct flow of execution. When evaluating the semantics of a behavior tree, we do not take into account any syntactic constraint.

As for the previous performance metrics, Alpaca is the least successful of all models. Values are low for all configurations. In the case of LlamaChat and CodeLlama, we observe an interesting behavior. When using a zero-shot prompt, performance increases when the model is fine-tuned. However, when adding an example in the prompt, performance decreases when using a fine-tuned model. This result shows again how impactful prompt design can be in guiding the LLM to achieve the correct solution. Nonetheless, semantic evaluation is performed only for the sake of completeness, since it disregards the syntactic correctness of the tree.

From this initial evaluation phase, we can conclude that in general, fine-tuned models perform better than base models in generating behavior trees in an XML format. They are faster and produce a correct solution more consistently, even with more complex tasks. The few exceptions are related to the use of examples in the prompt, and these situations are explored more in detail in the second evaluation phase. Additionally, we established that Alpaca, while achieving relatively good syntactic correctness when provided with examples, has poor performance across the board. Given these considerations, in the second phase, we will evaluate only LlamaChat and CodeLlama fine-tuned, both without and with examples.

\begin{table}[t]
\caption{Syntactic Correctness (phase 2)}
\label{syntax-table-phase2}
\centering
\begin{threeparttable}
\begin{tabular}{@{}llll@{}}
\toprule
& LlamaChat & CodeLlama    \\ \midrule
Zero-Shot & 88,9\% & 66,7\% \\
One-Shot & 88,9\% & 88,9\%\tnote{*} \\ \bottomrule
\end{tabular}
\begin{tablenotes}
\item[*] Note: max\_new\_tokens parameter limited to 1000, larger values increase accuracy
\end{tablenotes}
\end{threeparttable}
\end{table}

\begin{table}[t]
\caption{Validation (phase 2)}
\label{validation-table-phase2}
\centering
\begin{threeparttable}
\begin{tabular}{@{}lllllll@{}}
\toprule
       & \multicolumn{3}{l}{LlamaChat} & \multicolumn{3}{l}{CodeLlama} \\
       & ZS      & OS      & OS+SA     & ZS      & OS      & OS+SA     \\ \midrule
Task 1 & & \checkmark & \checkmark & & \checkmark & \checkmark \\
Task 2 & & \checkmark & \checkmark & & & \\
Task 3 & & & \checkmark & & \checkmark & \checkmark \\
Task 4 & & \checkmark & \checkmark & & \checkmark & \checkmark \\
Task 5 & & \checkmark & \checkmark & & & \\
Task 6 & &         & \checkmark &         &         &           \\
Task 7 & &         &           &         &         &           \\
Task 8 & &         & \checkmark &         &         &           \\
Task 9 & &         &           &         &         &    \\ \bottomrule      
\end{tabular}
\begin{tablenotes}
Success rate for behavior trees generated with zero-shot (ZS), and one-shot (OS) prompts. Additionally, the results of the corrected BTs are included (OS+SA).
\end{tablenotes}
\end{threeparttable}
\end{table}

\subsection{Validation}
During the second evaluation phase, we employed our custom-developed behavior trees validator to assess the overall correctness of the nine LLM-generated trees.
This validator is designed to examine and confirm the overall correctness of the generated behavior trees.
Additionally, further tests have been conducted within a simulation environment, in particular, we used the TurtleBot3 by ROBOTIS\footnote{https://emanual.robotis.com/docs/en/platform/turtlebot3/overview/} for navigation tasks leveraging the capabilities of Nav2.
For this reason, we developed a simple action client capable of sending navigation goals.
Each action specified in the previous tasks was wrapped within a corresponding node within the behavior tree.
With this approach, we aimed to ensure the reliability and functionality of the generated behavior trees in practical scenarios, both in simulated environments and real-world scenarios.

In this phase, we limited our analysis to the fine-tuned version of CodeLlama and LlamaChat. The two LLMs are prompted in the same way as the previous phase, however, we redesigned the examples in the one-shot prompts to be more in line with the target task. In practice, this means providing in the example at least one instance of each action being used to hint to the LLM the structure (i.e., name and type) of the parameters.

\begin{mdframed}[linecolor=light-gray]
\scriptsize{\textsf{\color{system}You will be provided a summary of a task performed by a robot, and your objective is to express this task as a behavior tree in XML format.}}
\vspace{0.6em}

    \noindent \scriptsize{\textsf{\color{prompt-example}The behavior tree represents a mobile robot tasked to visit two locations: (7,1) and (4,8). The available actions are: "MoveTo"}}
    \vspace{-0.6em}
    \begin{mdframed}[backgroundcolor=highlight-example, innertopmargin=-4pt, innerbottommargin=-4pt, innerleftmargin=2pt, innerrightmargin=2pt]
    \lstset{
        language=xml,
        stringstyle=\color{red},
        basicstyle=\scriptsize\ttfamily,
    }
\begin{lstlisting}
<root BTCPP_format="4">
    <BehaviorTree>
        <Sequence>
            <MoveTo x="7" y="1"/>
            <MoveTo x="4" y="8"/>
        </Sequence>
     </BehaviorTree>
</root>
\end{lstlisting}
    \end{mdframed}
    \vspace{-0.6em}

    \noindent \scriptsize{\textsf{\color{prompt-result}The behavior tree represents a mobile robot tasked to visit a sequence of locations: ((0,0), (2,3), (4, 7), (5, 11)). The available actions are: "MoveTo"}}
\end{mdframed}

For example, this is the prompt used for Task~1. On top, in grey, there is the system prompt used to define the general task of the LLM. Followed, in blue, by the instruction provided as an example. This is a simplified version of the target task where only two locations are visited. Then, the corresponding behavior tree where the format of the action is exemplified. Last, in green, there is the description of Task~1.

As before, first, we evaluate the syntactic correctness of the behavior trees generated by each model. Table~\ref{syntax-table-phase2} shows the results obtained. LlamaChat, both with zero-shot and one-shot prompts, is rather consistent and achieves syntactic correctness for most of the behavior trees. CodeLlama has problems obtaining correct results when no example is provided, possibly because, by being focused on code generation the model was exposed to more diverse types of behavior trees. Nonetheless, with a one-shot prompt, CodeLlama only fails to generate a syntactically correct behavior tree only in one case. It is worth noting that the failure is caused by reaching the maximum number of available tokens during generation. This problem appears when the LLM decomposes the behavior tree in multiple subtrees causing an unnecessarily lengthy solution. In general, multiple iterations of the same prompt can be used to achieve a syntactically correct result.

Table~\ref{validation-table-phase2} shows the results obtained after testing the behavior trees using our validation system. We performed the validation considering three different outputs. One is the direct output obtained by using zero-shot prompts, then the output of one-shot prompts as previously described, and lastly, the output of one-shot prompts corrected using a basic static analysis. The result of the static analysis on the behavior tree created using a zero-shot prompt is not reported since it provides no benefit.

\begin{mdframed}[linecolor=light-gray]
\lstset{
    language=diff
}
\begin{lstlisting}
  <root BTCPP_format="4">
  <BehaviorTree>
  <Sequence>
-   <Fallback>
      <Sequence>
-       <Action ID="moveToNewConfiguration"
-               goal_pose="{goal_pose}" />
+       <Action ID="moveToNewConfiguration" />
          <Sequence>
-           <Action ID="CheckForTarget"
-                   target="{target}" />
+           <Action ID="CheckForTarget" />
-             <Fallback>
                <Sequence>
-                 <Action ID="ApproachTarget"
-                         target="{target}" />
+                 <Action ID="ApproachTarget"/>
                </Sequence>
-               <Action ID="MoveBack" />
-             </Fallback>
          </Sequence>
      </Sequence>
-       <Action ID="MoveBack" />
-   </Fallback>
  </Sequence>
  </BehaviorTree>
  </root>
\end{lstlisting}
\end{mdframed}

This is an example of how behavior trees are corrected via static analysis. The corrected behavior trees are obtained by removing the extra parameters from the actions, and by removing unrecognized actions. In some cases, these changes lead to empty control sequences that are then removed.

The validation process shows that behavior trees generated with a zero-shot prompt consistently fail to achieve the target task. Most of the time this is because there is a mismatch between the provided parameters and the expected format. When provided with examples, both models obtain better results. Simpler tasks that contain basic execution flows are correctly generated. In more complex tasks that include a form of control flow (e.g., ``During navigation a location may become unreachable, if this happens, skip it''), the behavior tree fails because the LLM fails to understand the request or because actions are added to try to capture this control flow. In this second case, the behavior tree is successful after the action is removed via static analysis. In sum, the best results are achieved by LlamaChat when prompted with an example and after a cleanup of the behavior tree using static analysis. When inspecting the results, CodeLlama is more effective at creating complex and articulated behavior trees, often including subtrees, and comments. However, it fails to understand the request provided in the more complex prompts.

\section{Conclusion}
We introduce a novel approach in robotics, enabling large language models with at most 7 billion parameters to generate executable behavior trees for robots.
We compared Alpaca, LlamaChat, and CodeLlama, to identify the most suitable model for generating ready-to-use behavior trees. To enhance the generation capabilities of these base models, we created a new instruction-following dataset specifically designed for fine-tuning behavior tree generation.
The generated behavior trees have been evaluated in terms of syntactic accuracy, assessed with Groot2, and semantic accuracy, using a custom-developed validator.
Furthermore, their performance was assessed in simulation and real-world deployment on a physical robot. Our assessment shows how fine-tuned models perform better than base models. In particular,  LlamaChat is the best model overall, although CodeLlama still demonstrated notable performance. This work showcases the efficiency of compact large language models in directing autonomous agents like robots.
Finally, our approach is a plug-and-play solution, to transition from the LLM to direct execution on the robot.

The main limitation of using an LLM to generate robot behaviors is the challenging task of evaluating and validating the correctness of the solution. In our work we managed to do so using a custom-developed validator, however, it is limited to behaviors with a known solution. In the future, to deploy our system directly on the robot, we will focus on the automatic validation of the generated behavior tree. An option can be the use of milestones in the task to provide intermediate constraints and validate simpler behaviors. Alternatively, an auxiliary LLM can be used to regenerate the description from the XML and compare it with the original prompt.

The final aim of this work is to provide a direct interface between a user and the robot. Using this interface, the user can describe a task using natural language and the robot will execute it autonomously. Solving this challenge has significant potential to improve human-robot interaction and the capabilities of robots in various fields such as service robotics, industrial inspections, and last-mile logistics.

\section*{Acknowledgment}
This paper is supported by the FAIR (Future Artificial Intelligence Research) project, funded by the NextGenerationEU program within the PNRR-PE-AI scheme (M4C2, investment 1.3, line on Artificial Intelligence).

\bibliographystyle{plain}
\bibliography{bibliography}

\end{document}